\documentclass{article}

\usepackage{microtype}
\usepackage{graphicx}
\usepackage{subfigure}
\usepackage{booktabs}
\usepackage{tabularx}
\usepackage{adjustbox}

\usepackage{hyperref}

\usepackage[accepted]{mlsys2025}
\usepackage{amsmath}
\usepackage{amssymb}
\usepackage{algorithm}
\usepackage{algorithmic}  
\usepackage{caption}
\usepackage{bm}
\usepackage{xspace}
\usepackage{pifont}
\usepackage[table,x11names]{xcolor}
\usepackage{cleveref}
\usepackage{enumitem}
\usepackage{booktabs}     
\usepackage{multirow}      
\usepackage{graphicx}     
\usepackage{array}        
\usepackage{makecell}    
\usepackage{caption}      
\usepackage{adjustbox}

\newcommand{\sysname}{UniSparse\xspace}

\usepackage{multicol} 
\usepackage{multirow} 
\definecolor{MyHighlightColor}{RGB}{181,126,220}
\newcommand{\highlightrow}{\rowcolor{MyHighlightColor!15}}

\newcommand{\hc}[1]{\cellcolor{MyHighlightColor!15}#1}
\mlsystitlerunning{A Unified Sparse Attention via Multi-Granularity Compression}

\newcommand{\crunch}{\vspace{-1mm}}

\begin{document}

\twocolumn[
\mlsystitle{A Unified Sparse Attention via Multi-Granularity Compression}

\mlsyssetsymbol{equal}{*}

\begin{mlsysauthorlist}
\mlsysauthor{Siran Liu}{to,goo}
\mlsysauthor{Zane Cao}{goo}
\mlsysauthor{Yongchao He}{goo}
\end{mlsysauthorlist}

\mlsysaffiliation{to}{Peking University}
\mlsysaffiliation{goo}{SCITIX (SGP) TECH PTE. LTD}

\mlsyskeywords{Machine Learning, MLSys}

\vskip 0.3in

\begin{abstract}
Efficient long-context understanding and reasoning are increasingly vital for large language model (LLM) applications such as multi-turn dialogue and program analysis. However, the core \emph{self-attention} mechanism scales quadratically with sequence length, creating a fundamental computational bottleneck. Existing sparse attention methods alleviate this issue but face trade-offs: training-based methods are costly and cannot be directly applied as acceleration plugins for other models, while inference-time methods often compromise efficiency or cross-modal generality. To address these limitations, we present \sysname, a unified mechanism that introduces the notion of \emph{composite tokens}—compact representations that aggregate multi-granularity contextual information. Building on this abstraction, \sysname dynamically constructs sparse attention through \emph{multi-granularity compression} and \emph{block-level selection}, enabling efficient and hardware-friendly execution on GPU. Across multiple modalities and tasks ranging from synthetic benchmarks to real-world applications, \sysname consistently surpasses state-of-the-art sparse attention methods (e.g., MInference, XAttention, FlexPrefill) in both accuracy and efficiency, achieving $\ge$ 99\% of full-attention accuracy and up to 2.61$\times$ faster attention computation than FlashAttention.
\end{abstract}
]

\printAffiliationsAndNotice{} 

\section{Introduction}

The unprecedented success of Transformer-based~\cite{vaswani2017attention} large language models (LLMs) across natural language processing, computer vision, and multi-modal tasks is largely driven by their core \emph{self-attention} mechanism. This mechanism computes pairwise similarities between \emph{query} and \emph{key} token representations to produce contextualized outputs, forming a dense all-to-all interaction matrix among tokens. As applications increasingly demand understanding and generation over long contexts—ranging from multi-turn dialogue and document reasoning to code and program analysis—efficiently processing long sequences has become critical for both accuracy and usability. However, the computational cost of self-attention scales quadratically ($O(L^2)$) with the sequence length $L$, becoming the dominant bottleneck for long-context processing. For example, extending the context from 4K to 128K tokens—a trend in recent long-context models such as DeepSeek~\cite{deepseekapi}—increases the attention computation cost by about 1024$\times$, amounting to trillions of floating-point operations even with optimized kernels. While tensor, pipeline, or sequence parallelism~\cite{he2025sipipe, kim2025spd} can distribute memory across devices, they cannot mitigate the quadratic compute growth or the substantial inter-GPU communication it incurs. Consequently, dense attention remains computationally prohibitive for long-context modeling.

A variety of methods have been proposed to mitigate this quadratic overhead by introducing sparsity into self-attention, either learned during training or applied post-hoc at inference. 
Training-based approaches~\cite{yuan2025native, lu2025moba} learn data-dependent sparsity patterns that align well with model representations. 
In contrast, inference-time methods — including static~\cite{zaheer2020big,xiao2023efficient,he2025trianglemix} and dynamic ones~\cite{jiang2024minference,gao2024seerattention,xu2025xattention,lai2025flexprefill} — apply sparsification to pre-trained models.
To maximize GPU parallelism, these sparsification strategies employ a \emph{block-wise} design that enhances hardware efficiency by restricting computation to selected query–key block pairs indicated by a \emph{sparse mask matrix}~$\mathcal{M}$.

However, existing approaches expose a clear system tension between fidelity and efficiency. Training-coupled methods achieve accuracy but lack plug-and-play generality, while inference-time heuristics offer speed at the cost of robustness and adaptability (validated in \S\ref{sec:experiments}). In practice, this trade-off centers on how $\mathcal{M}$ is determined—typically via a lightweight \emph{proxy computation} that estimates block importance without performing full attention. Existing proxies either rely on expensive attention-like operations or on oversimplified heuristics that fail to capture semantic relevance, leading to a difficult trade-off between proxy accuracy and cost. This persistent gap raises a question: \textit{is there a sparse attention mechanism that is both hardware-efficient and robust, generalizing across modalities without retraining?}

This paper presents \sysname, a unified, hardware-friendly mechanism for \emph{dynamic sparse attention} that operates at inference time and exhibits strong generalization across different modalities. The key insight underlying \sysname is that \emph{semantically meaningful attention patterns can be faithfully inferred within a drastically compressed token space}. Instead of evaluating token importance at full sequence resolution, \sysname constructs \emph{composite tokens}—coarse-grained summaries formed through spatial aggregation of fine-grained tokens—that preserve essential contextual structure while reducing proxy computation cost.

Building on this insight, \sysname implements a \emph{multi-granularity compression mechanism} that projects query and key representations into a reduced space along both sequence and head dimensions, enabling lightweight yet expressive proxy computations. A \emph{dynamic block selection algorithm} then computes attention scores in the compressed space, aggregates them to block-level importance, and identifies the most salient region. The resulting mask $\mathcal{M}$ guides efficient block-sparse attention computation using standard attention kernels, e.g., FlashAttention~\cite{dao2022flashattention}, ensuring both software portability and hardware efficiency.

Together, these techniques enable \sysname to deliver high-fidelity sparse attention without any model retraining or task-specific heuristics. Across text and multi-modal benchmarks, \sysname consistently  
surpasses state-of-the-art sparse methods in both attention score accuracy and computational efficiency, and does so using 1.5--2$\times$ fewer attention blocks. End-to-end, \sysname attains up to 2.61$\times$ faster attention computation with over 99\% accuracy retention, demonstrating that its composite tokens, together with multi-granularity compression and block-level selection, generalize effectively across modalities and model architectures while preserving full-attention quality.

In summary, this paper makes the following contributions: 
(1) We characterize the trade-off between attention accuracy and computational cost in existing sparse attention methods, highlighting its system-level implications; 
(2) We propose \sysname, which realizes efficient sparse attention by introducing composite tokens for compact multi-granularity context representation, enabled through multi-granularity compression and dynamic block selection;
and (3) We implement and evaluate \sysname across diverse workloads, achieving up to 2.61$\times$ speedup in attention computation with $\le$1\% accuracy loss.

\section{Background}

\label{sec:Background}

\subsection{LLM Inference}

Modern LLMs rely on stacked Transformer blocks combining attention and MLP layers. During inference, the \emph{prefill phase}—a full-sequence forward pass with quadratic attention cost—dominates runtime in long-context tasks (e.g., code analysis, document understanding). This work targets prefill sparsity optimization, as the subsequent \emph{decoding phase} is comparatively lightweight.

To see this bottleneck concretely, consider a single attention head for a sequence of length $L$ and per-head dimension $d_k$. Let $\mathbf{Q}, \mathbf{K}, \mathbf{V} \in \mathbb{R}^{L \times d_k}$ be the query, key, and value matrices obtained via learned projections. The attention output $\mathbf{O} \in \mathbb{R}^{L \times d_k}$ is $\mathbf{O} = \text{softmax}\Big(\frac{\mathbf{Q}\mathbf{K}^T}{\sqrt{d_k}}\Big)\mathbf{V}.$
The dominant computation is forming the $L \times L$ attention score matrix $\mathbf{Q}\mathbf{K}^T$, giving complexity $\mathcal{O}(L^2 d_k)$.  

Modern attention implementations on GPU, such as FlashAttention~\cite{dao2022flashattention}, compute attention in a block-wise manner. The sequence is partitioned into blocks of size $S$, producing $N=L/S$ query blocks $\{\mathbf{Q}_i\}_{i=1}^N$ and key blocks $\{\mathbf{K}_j\}_{j=1}^N$. Attention is computed tile-by-tile, loading one query block and one key block at a time into fast on-chip memory. This improves GPU utilization and memory efficiency for long sequences. 

\subsection{Block-Sparse Attention}

While block-wise implementations improve GPU throughput, the quadratic complexity $\mathcal{O}(L^2 d_k)$ still dominates for very long sequences. Block-sparse attention~\cite{jiang2024minference, guo2024blocksparse} addresses this by computing attention only on a subset of block pairs: A block-level binary mask $\mathcal{M} \in \{0,1\}^{N \times N}$ specifies which query-key block pairs to compute, where $\mathcal{M}_{ij} = 1$ indicates that query block $i$ attends to key block $j$, and $\mathcal{M}_{ij} = 0$ skips that pair. For auto-regressive LLMs, causality requires $\mathcal{M}_{ij} = 0$, $\forall j>i$.

The attention output is computed by masking the attention scores before softmax: $\mathbf{O} = \text{softmax}\Big(\frac{\mathbf{Q}\mathbf{K}^T}{\sqrt{d_k}} + \tilde{\mathcal{M}}\Big)\mathbf{V}$,
where $\tilde{\mathcal{M}}$ is the token-level mask derived from $\mathcal{M}$, with $0$ for selected blocks and $-\infty$ for masked positions.

The computational savings are determined by the sparsity ratio $\rho = 1 - \frac{2}{N(N+1)}\sum_{i \geq j}\mathcal{M}_{ij}$ under causality. The total cost of block-sparse attention can be decomposed as
\begin{equation}
    \label{eq:total_cost}
    \setlength{\abovedisplayskip}{2.5pt}
    \setlength{\belowdisplayskip}{2.5pt}
    C_{\text{total}} = \underbrace{C_{\text{select}}}_{\text{Mask Determination}} + \underbrace{\rho \cdot C_{\text{full}}}_{\text{Sparse Computation}}
\end{equation}
where $C_{\text{full}} = \mathcal{O}(L^2 d_k)$ is the cost of full attention on selected blocks, and $C_{\text{select}}$ is the overhead of determining $\mathcal{M}$. For a given sparsity $\rho$, most methods have similar sparse computation costs, so $C_{\text{select}}$ becomes the main factor distinguishing their efficiency. The fundamental challenge is \emph{designing a selection mechanism that balances low $C_{\text{select}}$ with high accuracy in identifying important blocks, i.e., $\mathcal{M}$}.

\section{Related Work}
\label{sec:related_work}

To reduce the quadratic cost of attention, numerous methods exploit sparse patterns. These approaches differ along two axes: \textit{when} sparsity is applied—training or inference—and \textit{how} it is defined—static versus dynamic. We review these methods and highlight their trade-offs in adaptivity, overhead, and generality.

\subsection{Training-Based Sparse Attention}
Training-based sparsity methods embed sparse structures directly into the model during pre-training or fine-tuning. For example, DeepSeek’s Native Sparse Attention employs a hierarchical strategy to learn sparse patterns~\cite{yuan2025native}, while Kimi’s Mixture-of-Block-Attention trains a lightweight router to select relevant blocks~\cite{lu2025moba}. Although such training-based sparsity enables data-aligned attention patterns, the additional training overhead and the tight coupling between sparse structures and model weights hinder flexibility—these approaches cannot serve as drop-in accelerators for existing densely-trained models. Moreover, their learned patterns are often domain-specific: a router trained on text rarely generalizes to video, necessitating expensive modality-specific retraining. Hence, real-world deployments favor \textit{inference-time} sparsification techniques.

\subsection{Inference-Time Sparse Attention}
\subsubsection{Inference-Time Static Attention}
Static methods apply a pre-defined, input-agnostic sparse pattern during inference, incurring zero selection overhead ($C_{\text{select}} = 0$). Examples include BigBird~\cite{zaheer2020big}, which combines sliding window, global, and random attention; StreamLLM~\cite{xiao2023efficient}, which retains early sink tokens; and TriangleMix~\cite{he2025trianglemix}, which skips intermediate Q--K block interactions in deeper layers.
While computationally efficient, such fixed patterns lack adaptability to input-dependent long-range dependencies or modality-specific structures. Originally tuned for text, most static designs struggle to capture spatio-temporal or cross-modal relationships in domains like video or audio, leading to degraded comprehension and reasoning.

\subsubsection{Inference-Time Dynamic Attention}
Dynamic methods construct the mask $\mathcal{M}$ on-the-fly via lightweight proxy computation, with the central challenge of balancing proxy accuracy (model quality), computational cost, and hardware efficiency. Existing approaches can be broadly grouped into two families.

\noindent
\emph{(1) High-Fidelity Proxy Methods.}
These methods aim to closely approximate dense attention patterns with more expressive but costly proxies. 
For instance, MInference~\cite{jiang2024minference} conducts offline searches to assign predefined patterns, while SeerAttention~\cite{gao2024seerattention} introduces learnable gating networks for runtime selection.  
Although effective, such designs are heavyweight: they require model-specific pre-processing or additional training, which compromises generality and plug-and-play usability.

\noindent
\emph{(2) Low-Complexity Proxy Methods.}
An alternative line of work employs simple heuristics to minimize runtime overhead. 
XAttention~\cite{xu2025xattention}, for example, uses the sum of anti-diagonal elements within each block as a low-cost proxy, while FlexPrefill~\cite{lai2025flexprefill} exploits the attention pattern of the last query block as a probe to select important vertical and slash structures. 
These approaches achieve high throughput, but their effectiveness hinges on the alignment between the heuristic and the data modality. 
In complex multimodal tasks, the assumption that a single block can represent global context often fails, leading to sub-optimal sparsity and accuracy degradation.

\subsection{Discussion and Open Question}
\label{sec:summary_gap}
Overall, existing inference-time dynamic approaches exhibit a persistent trade-off. High-fidelity proxy methods can achieve high accuracy but often lack practicality and generality due to their complexity and the need for model-specific preprocessing. In contrast, low-cost static heuristic approaches are computationally efficient but fragile, as they rely on oversimplified, modality-specific assumptions. 
This tension motivates a fundamental question guiding our work: \textit{Is there a proxy computation paradigm that is structurally simple and hardware-friendly, yet robust enough to generate effective sparse patterns across diverse modalities—without any model training or adaptation?}

\begin{figure}[!t] 
\setlength{\abovecaptionskip}{-1pt}
\setlength{\belowcaptionskip}{-2pt}
  \centering
  \includegraphics[width=0.95\columnwidth]{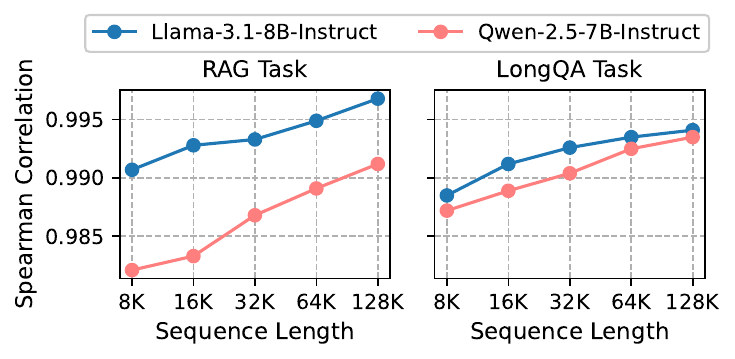}
  \caption{Spearman correlation between compressed and original block importance rankings on HELMET.}
  \label{fig:pre}
  \crunch
  \crunch
  \crunch
\end{figure}

\section{Design}
This section introduces \sysname, a unified mechanism for dynamic sparse attention. We begin with the core intuition (\S\ref{subsec:intuition}), followed by a description of the compression mechanism (\S\ref{subsec:compression}) and the dynamic block selection algorithm (\S\ref{subsec:selection}), and conclude with a complexity analysis (\S\ref{subsec:complexity}).

\begin{figure*}[!t]  
  \centering
  \setlength{\abovecaptionskip}{-1pt}
  \setlength{\belowcaptionskip}{-6pt}
  \includegraphics[width=\textwidth]{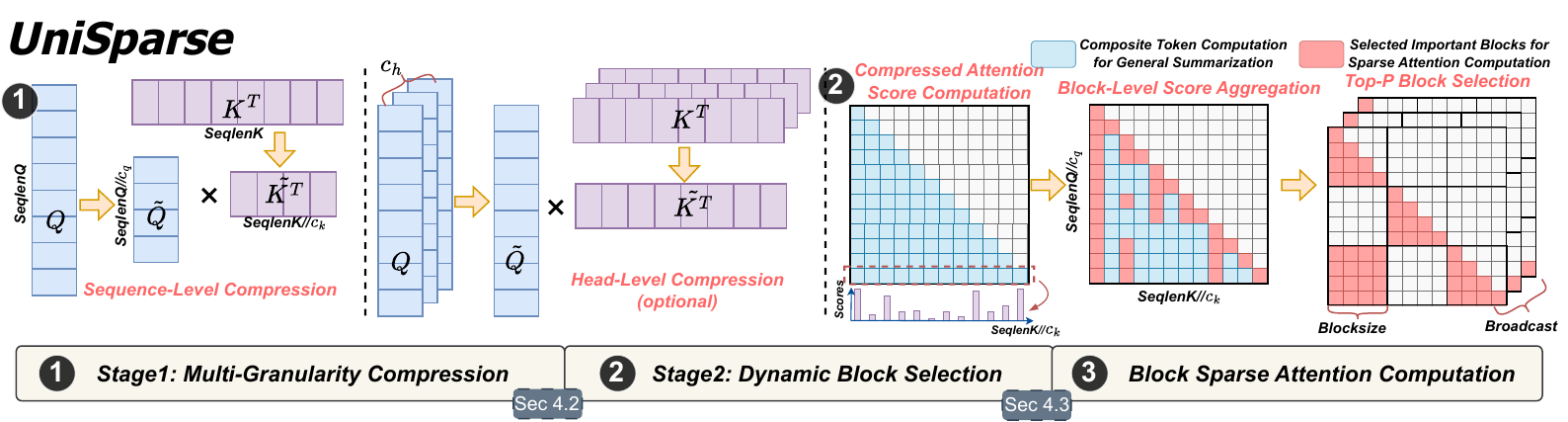}
  \caption{Workflow of \sysname for efficient sparse attention.}
  \label{fig:workflow}
  \crunch
  \crunch
  \crunch
\end{figure*}

\subsection{Intuition and Overview}
\label{subsec:intuition}

\textbf{Key Insight: Composite Tokens as Universal Summaries.} The central hypothesis of \sysname is that \textit{semantically meaningful attention patterns can be reliably approximated in a drastically compressed token space}. Rather than computing importance scores at the native token granularity or relying on partial evaluations that sample only a subset of query-key interactions, we propose to construct \emph{\textbf{composite tokens}}---coarse-grained summary vectors obtained through spatial pooling (i.e., aggregating neighboring tokens within local regions) of fine-grained tokens. The intuition is rooted in the observation that tokens within local neighborhoods often exhibit semantic coherence, making their aggregate representation a faithful proxy for the collective importance of the underlying fine-grained tokens.

Formally, consider a query block $\mathbf{Q}_i \in \mathbb{R}^{S \times d_k}$ and key block $\mathbf{K}_j \in \mathbb{R}^{S \times d_k}$. Instead of computing the full $S \times S$ attention score matrix between them (which would scale the selection cost to be comparable to the attention computation itself), we compress them into composite representations $\tilde{\mathbf{Q}}_i \in \mathbb{R}^{S' \times d_k}$ and $\tilde{\mathbf{K}}_j \in \mathbb{R}^{S' \times d_k}$ where $S' \ll S$, compute a much smaller $S' \times S'$ score matrix, and use block-wise aggregated scores from this compressed space to determine block importance.

To validate this hypothesis, we conduct a preliminary experiment on the HELMET~\cite{yen2025helmet}, measuring the Spearman correlation between block importance rankings computed in compressed space (compression ratio $c=8$) versus original token-level rankings. Figure~\ref{fig:pre} shows consistently high correlation ($\rho > 0.98$) across different sequence lengths on both RAG and LongQA tasks for Llama and Qwen models, while reducing the proxy computation cost by orders of magnitude. This strong rank preservation validates our core insight: block selection fundamentally requires accurate \textit{relative ordering} rather than absolute scores, as we elaborate in the following analysis (\S\ref{subsec:compression}).

Crucially, this compression-based proxy is \textit{modality-agnostic}. Unlike methods that rely on domain-specific heuristics---such as structural assumptions about sequential dependencies or partial query-key evaluations designed for specific data patterns---spatial pooling is a universal operation applicable to any sequential data, be it text tokens, video frame patches, or audio segments. This universality is the foundation of \sysname's cross-modality effectiveness.

\textbf{Overview.}
\sysname addresses the two fundamental challenges of dynamic sparse attention: (\textit{i}) how to obtain a compact yet faithful representation of attention structure, and (\textit{ii}) how to identify the most important regions for computation. 
Accordingly, as shown in Figure~\ref{fig:workflow} it operates in two stages. 
In the \emph{multi-granularity compression} stage (\ding{182}), the query and key matrices are independently compressed along the sequence dimension (and optionally the head dimension) using average pooling, transforming fine-grained tokens into coarser composite tokens. 
In the subsequent \emph{dynamic block selection} stage (\ding{183}), attention scores are computed in the compressed token space, aggregated at the block level, and filtered through a Top-$P$ mechanism to identify the most important blocks. 
The resulting sparse mask $\mathcal{M}$ is then used to perform efficient block-sparse attention with standard attention kernels such as FlashAttention (\ding{184}).

\subsection{Multi-Granularity Compression}
\label{subsec:compression}

\textbf{Sequence-Level Compression.} Given a sequence of length $L$ with $N = L/S$ blocks of size $S$, we define two compression factors: $c_q$ for queries and $c_k$ for keys, where $c_q, c_k \in \mathbb{Z}^+$. The compressed sequence lengths are $L'_q = L/c_q$ and $L'_k = L/c_k$, yielding $N'_q = L'_q/S$ and $N'_k = L'_k/S$ compressed blocks.
For the $i$-th query block $\mathbf{Q}_i = [\mathbf{q}_{i,1}, \ldots, \mathbf{q}_{i,S}]^T \in \mathbb{R}^{S \times d_k}$, where $\mathbf{q}_{i,t} \in \mathbb{R}^{d_k}$ is the $t$-th token within the block, we construct compressed block $\tilde{\mathbf{Q}}_{i'} \in \mathbb{R}^{S/c_q \times d_k}$ through average pooling:
\begin{equation}
    \setlength{\abovedisplayskip}{3pt}
    \setlength{\belowdisplayskip}{3pt}
    \label{eq:avgq}
    \tilde{\mathbf{q}}_{i',t'} = \frac{1}{c_q} \sum_{m=0}^{c_q-1} \mathbf{q}_{i, t' \cdot c_q + m}, \quad t' = 1, 2, \ldots, S/c_q
\end{equation}

where $i' = \lfloor i / (c_q/S) \rfloor$ maps the original block index to the compressed space. The compressed query matrix is $\tilde{\mathbf{Q}} \in \mathbb{R}^{L'_q \times d_k}$. Similarly, we obtain compressed key matrix $\tilde{\mathbf{K}} \in \mathbb{R}^{L'_k \times d_k}$ using compression factor $c_k$.

\textbf{Head-Level Compression (Optional).} To further reduce overhead, we introduce an optional head dimension compression with factor $c_h$. For a multi-head attention layer with $H$ heads, we group every $c_h$ consecutive heads and compute their average:
\begin{equation}
    \setlength{\abovedisplayskip}{3pt}
    \setlength{\belowdisplayskip}{3pt}
    \tilde{\mathbf{Q}}^{(h')} = \frac{1}{c_h} \sum_{m=0}^{c_h-1} \tilde{\mathbf{Q}}^{(h' \cdot c_h + m)}, \quad h' = 0, 1, \ldots, \frac{H}{c_h} - 1
\end{equation}
This yields $H' = H/c_h$ compressed heads. The same operation is applied to keys. Critically, the sparse pattern determined from compressed heads is \textit{broadcast} to all original heads within each group, maintaining per-head expressiveness while amortizing the selection cost. Head compression can be viewed as a distillation of consensus importance signals across heads, with effectiveness varying across different tasks and model architectures. 

\definecolor{comment_color_2}{RGB}{64,128,128}
\newcommand{\LineComment}[1]{\vspace*{0.5em}\small\textcolor{comment_color_2}{\textit{\# #1}}}

\begin{algorithm}[t]
\captionsetup[algorithm]{singlelinecheck=off}
\caption{\sysname: Multi-Granularity Block Selection}
\label{alg:unisparse}
\begin{algorithmic}
\STATE {\bfseries Input:} $\boldsymbol{Q}, \boldsymbol{K}, \boldsymbol{V} \in \mathbb{R}^{L \times d_k}$, block size $S$, compression factors $c_q, c_k, c_h$, threshold $P \in (0,1]$

\LineComment{Step 1: Multi-Granularity Compression}
\STATE $\tilde{\boldsymbol{Q}}, \tilde{\boldsymbol{K}} \gets \texttt{AvgPool}_{\text{seq}}(\boldsymbol{Q}, c_q), \texttt{AvgPool}_{\text{seq}}(\boldsymbol{K}, c_k)$

\IF{$c_h > 1$}
    \STATE $\tilde{\boldsymbol{Q}}, \tilde{\boldsymbol{K}} \gets \texttt{AvgPool}_{\text{head}}(\tilde{\boldsymbol{Q}}, c_h), \texttt{AvgPool}_{\text{head}}(\tilde{\boldsymbol{K}}, c_h)$
\ENDIF

\LineComment{Step 2: Compressed Attention Computation}
\STATE $\tilde{\boldsymbol{P}} \gets \tilde{\boldsymbol{Q}} \tilde{\boldsymbol{K}}^T / \sqrt{d_k}$
\STATE $\tilde{\boldsymbol{A}} \gets \texttt{softmax}(\tilde{\boldsymbol{P}})$

\LineComment{Step 3: Block-Level Score Aggregation}
\STATE $\boldsymbol{Score} \gets \texttt{BlockAggregate}(\tilde{\boldsymbol{A}}, S, c_q, c_k)$

\LineComment{Step 4: Top-$P$ Block Selection with Causal Mask}
\FOR{$i = 1$ to $N$}
    \STATE $\boldsymbol{Score}[i, j] \gets -\infty$ for all $j > i$
    \STATE $\mathcal{T}_P(i) \gets \texttt{TopP\_Select}(\boldsymbol{Score}[i, :], P)$
    \STATE $\boldsymbol{\mathcal{M}}[i, \mathcal{T}_P(i)] \gets 1$
\ENDFOR

\IF{$c_h > 1$}
    \STATE $\boldsymbol{\mathcal{M}} \gets \texttt{Broadcast}_{\text{head}}(\boldsymbol{\mathcal{M}}, c_h)$
\ENDIF

\LineComment{Step 5: Block-Sparse Attention Computation}
\STATE $\boldsymbol{O} \gets \texttt{SparseFlashAttention}(\boldsymbol{Q}, \boldsymbol{K}, \boldsymbol{V}, \boldsymbol{\mathcal{M}})$

\STATE {\bfseries Output:} $\boldsymbol{O} \in \mathbb{R}^{L \times d_k}$
\end{algorithmic}
\end{algorithm}
  \crunch
  \crunch
  \crunch

\textbf{Why Compression Preserves Importance.} 
The effectiveness of compression in block importance estimation stems from a key insight: block selection is essentially a \textit{ranking} rather than a scoring problem. As long as compression preserves the relative ordering of block importance, selection quality remains high. Average pooling over composite tokens naturally maintains this ordering by summarizing local semantics into representative vectors that reflect the collective relevance of fine-grained tokens. Our experiments (\S\ref{sec:experiments}) confirm that such compression-based rankings closely match those derived from full token-level computations.

\subsection{Dynamic Block Selection}
\label{subsec:selection}

\textbf{Compressed Attention Score Computation.} Using the compressed queries $\tilde{\mathbf{Q}} \in \mathbb{R}^{L'_q \times d_k}$ and keys $\tilde{\mathbf{K}} \in \mathbb{R}^{L'_k \times d_k}$, we compute the attention score matrix in the compressed space $\tilde{\mathbf{P}} = \frac{\tilde{\mathbf{Q}} \tilde{\mathbf{K}}^T}{\sqrt{d_k}} \in \mathbb{R}^{L'_q \times L'_k}$.
This matrix has size $(L/c_q) \times (L/c_k)$, which is a factor of $c_q \cdot c_k$ smaller than the full $L \times L$ score matrix. We then apply \texttt{softmax} normalization at the composite token granularity $\tilde{\mathbf{A}} = \texttt{softmax}(\tilde{\mathbf{P}}) \in \mathbb{R}^{L'_q \times L'_k}$,
where \texttt{softmax} is applied row-wise across all composite key tokens for each composite query token. This compressed attention matrix, operating at the coarse-grained composite token level, serves as our lightweight proxy for block importance estimation.

\textbf{Block-Level Score Aggregation.} To determine which blocks in the \textit{original} space are important, we aggregate the compressed attention scores at block granularity. For each query block $\mathbf{Q}_i$ (in the original space), we identify its corresponding region in the compressed space and sum the attention scores to each key block $\mathbf{K}_j$:
\begin{equation}
    \setlength{\abovedisplayskip}{3pt}
    \setlength{\belowdisplayskip}{3pt}
    \texttt{Score}(i, j) = \sum_{t' \in \mathcal{R}_q(i)} \sum_{s' \in \mathcal{R}_k(j)} \tilde{\mathbf{A}}_{t', s'}
\end{equation}
where $\mathcal{R}_q(i)$ denotes the set of compressed token indices corresponding to original query block $i$, and $\mathcal{R}_k(j)$ for key block $j$. Specifically:
\begin{align}
    \setlength{\abovedisplayskip}{3pt}
    \setlength{\belowdisplayskip}{3pt}
    \mathcal{R}_x(u) = \left\{ z' \mid 
    \left\lfloor \frac{z' \cdot c_x}{S} \right\rfloor = u \right\},
    \quad x \in \{q, k\}
\end{align}
where $u = i$ and $z' = t'$ when $x = q$, and $u = j$, $z' = s'$ when $x = k$.
This aggregation produces a block-level score matrix $\mathbf{Score} \in \mathbb{R}^{N \times N}$, where $N = L/S$ is the number of blocks in the original space.

\textbf{Top-$P$ Block Selection.} For each query block $i$, we select the top-$P$ key blocks based on their aggregated scores. The sparsity is controlled by a threshold $P \in (0, 1]$, representing the proportion of total attention mass to retain. Formally, we define the mask entry as $\mathcal{M}_{ij} = 1$ if $j \in \mathcal{T}_P(i)$ and $0$ otherwise, 
where $\mathcal{T}_P(i)$ is the set of key block indices selected for query block $i$. We construct $\mathcal{T}_P(i)$ by sorting key blocks in descending order of $\texttt{Score}(i, j)$ and greedily accumulating blocks until the cumulative score reaches the threshold:
{\fontsize{9pt}{14pt}\selectfont
\begin{equation}
    \setlength{\abovedisplayskip}{3pt}
    \setlength{\belowdisplayskip}{3pt}
    \mathcal{T}_P(i) = \left\{ j \mid \frac{\sum_{j' \in \mathcal{T}_P(i)} \texttt{Score}(i, j')}{\sum_{j'=1}^{N} \texttt{Score}(i, j')} \geq P \right\}
\end{equation}}
Finally, we summarize the overall workflow of \sysname in Algorithm~\ref{alg:unisparse}.

\subsection{Complexity Analysis}
\label{subsec:complexity}

\textbf{Selection Overhead.} Understanding the computational cost of dynamic sparsity is critical for evaluating its practical efficiency. In particular, the overhead introduced by block selection often determines whether a sparse attention method offers real runtime benefits. 
We therefore analyze the computational complexity of \sysname (Algorithm~\ref{alg:unisparse}), focusing on the selection overhead $C_{\text{select}}$---the primary distinguishing factor among different sparse attention methods. The total overhead decomposes as:

\begin{equation}
\resizebox{\columnwidth}{!}{
    $
    \underbrace{\mathcal{O}(Lhd_k)}_{\texttt{Compression}} + \underbrace{\mathcal{O}\left(\frac{L^2 h d_k}{c_q \cdot c_k \cdot c_h}\right)}_{\text{\textbf{\texttt{Compressed QK}}}} + \underbrace{\mathcal{O}\left(\frac{L^2 h}{c_q \cdot c_k \cdot c_h}\right)}_{\texttt{softmax} \& \texttt{Aggregation}} + \underbrace{\mathcal{O}(N^2 h \log N)}_{\texttt{Top-}P}
    $
    }
\end{equation}
where $h$ is the number of attention heads. The compressed attention score computation dominates, with complexity $\mathcal{O}(L^2 h d_k / (c_q \cdot c_k \cdot c_h))$. The \texttt{softmax} and \texttt{aggregation} have the same order of complexity but with lower constants. The \texttt{Top-$P$ selection} has complexity $\mathcal{O}(N^2 h \log N) = \mathcal{O}((L^2 h / S^2) \log(L/S))$, which is negligible for reasonably large block sizes (e.g., $S = 128$).

\begin{table}[t]
    \centering
    \caption{Complexity and evaluation scope of dynamic sparse attention methods.}
    \label{tab:complexity_comparison}
    \resizebox{1\columnwidth}{!}{
    \begin{tabular}{l|c|c}
        \toprule
        \textbf{Method} & \textbf{Selection Complexity $C_{\text{select}}$} & \textbf{Evaluation Scope} \\
        \midrule
        Minference & $\mathcal{O}(S \cdot L \cdot h \cdot d_k) + \mathcal{O}_{\text{offline}}$ & \textit{Local(Last Query Block)} \\
        FlexPrefill & $\mathcal{O}(S \cdot L \cdot h \cdot d_k)$ & \textit{Local(Last Query Block)} \\
        XAttention & $\mathcal{O}(L^2 h d_k / \text{stride})$ & \textit{Global(Strided Sampling)} \\
        \textbf{\sysname} & $\mathcal{O}(L^2 h d_k / (c_q c_k c_h))$ & \textit{Global(Compressed Space)} \\
        \bottomrule
    \end{tabular}
    }
\end{table}

\begin{table*}[!t]
    \renewcommand{\arraystretch}{0.9}
    \centering
    \caption{Performance comparison on RULER across context lengths from 4K to 128K tokens. We report accuracy scores at each length and average sparsity ratios. The best results among sparse methods at similar sparsity levels are highlighted.}
    \label{tab:ruler}
    \setlength{\aboverulesep}{1.5pt}
    \setlength{\belowrulesep}{1.5pt}
    \setlength{\heavyrulewidth}{1.5pt}
    \resizebox{0.85\textwidth}{!}{
    \begin{tabular}{l|l|c|cccccc|c}
        \toprule
        \textbf{Model} & \textbf{Method} & \textbf{Sparsity (↑)} & \textbf{4K} & \textbf{8K} & \textbf{16K} & \textbf{32K} & \textbf{64K} & \textbf{128K} & \textbf{Overall (↑)} \\
        \midrule
        \multirow{9}{*}{\rotatebox{90}{\textit{Llama-3.1-8B-Instruct}}} 
        & FlashAttention & - & 97.30 & 96.95 & 96.42 & 93.96 & 89.96 & 82.65 & 92.87 \\
        \cmidrule{2-10}
        & FlexPrefill (0.95) & 64.16\% & 95.58 & 96.29 & 95.79 & \textbf{94.38} & \textbf{90.89} & \textbf{81.31} & 92.37 \\
        & XAttention (0.9) & 56.44\% & 97.21 & 96.57 & 95.85 & 93.73 & 89.02 & 79.17 & 91.93 \\
        & \hc{\textbf{UniSparse (0.9)}} & \hc{58.90\%} & \hc{\textbf{97.38}} & \hc{\textbf{96.59}} & \hc{\textbf{96.43}} & \hc{94.25} & \hc{89.29} & \hc{80.73} & \hc{\textbf{92.44}} \\
        \cmidrule{2-10}
        & MInference & 50.39\% & 97.29 & 96.69 & 96.71 & 93.40 & 88.67 & 81.38 & 92.35 \\
        & FlexPrefill (0.99) & 36.51\% & \textbf{97.33} & \textbf{96.79} & \textbf{96.35} & 93.99 & \textbf{90.53} & 80.52 & \textbf{92.59} \\
        & XAttention (0.95) & 44.33\% & 97.27 & 96.49 & 96.15 & 93.82 & 90.16 & 81.34 & 92.54 \\
        & \hc{\textbf{UniSparse (0.95)}} & \hc{46.33\%} & \hc{97.16} & \hc{96.52} & \hc{96.31} & \hc{\textbf{94.03}} & \hc{89.40} & \hc{\textbf{81.94}} & \hc{92.56} \\
        \midrule
        \multirow{9}{*}{\rotatebox{90}{\textit{Qwen2.5-7B-Instruct}}} 
        & FlashAttention & - & 94.77 & 89.42 & 87.84 & 85.46 & 81.18 & 70.92 & 84.93 \\
        \cmidrule{2-10}
        & FlexPrefill (0.95) & 64.14\% & 83.78 & 81.45 & 80.20 & 76.50 & 71.21 & 58.06 & 75.20 \\
        & XAttention (0.9) & 58.17\% & 92.84 & 87.91 & 85.02 & 82.80 & 77.14 & 66.58 & 82.05 \\
        & \hc{\textbf{UniSparse (0.9)}} & \hc{60.12\%} & \hc{\textbf{93.23}} & \hc{\textbf{87.69}} & \hc{\textbf{85.99}} & \hc{\textbf{84.26}} & \hc{\textbf{79.25}} & \hc{\textbf{68.00}} & \hc{\textbf{83.07}} \\
        \cmidrule{2-10}
        & MInference & 39.32\% & \textbf{94.68} & \textbf{89.65} & \textbf{88.56} & 85.04 & 78.47 & 65.08 & 83.58 \\
        & FlexPrefill (0.99) & 37.75\% & 92.42 & 88.77 & 85.91 & 84.13 & 77.89 & 67.69 & 82.80 \\
        & XAttention (0.95) & 46.44\% & 93.87 & 88.36 & 86.84 & 84.15 & 79.04 & \textbf{69.51} & 83.63 \\
        & \hc{\textbf{UniSparse (0.95)}} & \hc{47.25\%} & \hc{94.31} & \hc{88.96} & \hc{87.19} & \hc{\textbf{85.23}} & \hc{\textbf{80.42}} & \hc{69.47} & \hc{\textbf{84.26}} \\
        \bottomrule
    \end{tabular}
    }
\end{table*}

\begin{table*}[!t]
    \renewcommand{\arraystretch}{0.9}
    \centering
    \caption{Performance comparison on HELMET across context lengths from 8K to 128K tokens (metrics follows Table~\ref{tab:ruler}).}
    \label{tab:helmet}
    \setlength{\aboverulesep}{1.5pt}
    \setlength{\belowrulesep}{1.5pt}
    \setlength{\heavyrulewidth}{1.5pt}
    \resizebox{0.77\textwidth}{!}{
    \begin{tabular}{l|l|c|ccccc|c}
        \toprule
        \textbf{Model} & \textbf{Method} & \textbf{Sparsity (↑)} & \textbf{8K} & \textbf{16K} & \textbf{32K} & \textbf{64K} & \textbf{128K} & \textbf{Overall(↑)} \\
        \midrule
        \multirow{9}{*}{\rotatebox{90}{\textit{Llama-3.1-8B-Instruct}}} 
        & FlashAttention & - & 61.06 & 58.73 & 56.57 & 55.92 & 49.79 & 56.41 \\
        \cmidrule{2-9}
        & FlexPrefill (0.95) & 65.92\% & 55.89 & 54.73 & 53.14 & 50.45 & 48.42 & 52.53 \\
        & XAttention (0.9) & 59.80\% & 59.81 & 57.40 & 55.55 & 53.74 & 47.74 & 54.85 \\
        & \hc{\textbf{UniSparse (0.9)}} & \hc{61.82\%} & \hc{\textbf{60.86}} & \hc{\textbf{57.70}} & \hc{\textbf{56.21}} & \hc{\textbf{54.24}} & \hc{\textbf{48.91}} & \hc{\textbf{55.58}} \\
        \cmidrule{2-9}
        & MInference & 56.34\% & \textbf{61.44} & 57.96 & 55.49 & 53.69 & 48.29 & 55.37 \\
        & FlexPrefill (0.99) & 38.52\% & 60.66 & 58.05 & \textbf{56.22} & \textbf{55.44} & 49.68 & 56.01 \\
        & XAttention (0.95) & 47.45\% & 60.76 & 57.80 & 55.92 & 54.75 & 49.37 & 55.72 \\
        & \hc{\textbf{UniSparse (0.95)}} & \hc{48.65\%} & \hc{61.37} & \hc{\textbf{58.34}} & \hc{\textbf{56.22}} & \hc{55.27} & \hc{\textbf{50.22}} & \hc{\textbf{56.21}} \\
        \midrule
        \multirow{9}{*}{\rotatebox{90}{\textit{Qwen2.5-7B-Instruct}}} 
        & FlashAttention & - & 55.10 & 50.54 & 48.51 & 44.85 & 39.33 & 47.67 \\
        \cmidrule{2-9}
        & FlexPrefill (0.95) & 63.03\% & 44.23 & 40.79 & 39.87 & 36.28 & 30.06 & 38.25 \\
        & XAttention (0.9) & 59.08\% & 52.46 & 47.99 & 45.63 & 43.17 & 36.92 & 45.23 \\
        & \hc{\textbf{UniSparse (0.9)}} & \hc{60.84\%} & \hc{\textbf{53.13}} & \hc{\textbf{48.62}} & \hc{\textbf{46.93}} & \hc{\textbf{44.22}} & \hc{\textbf{38.11}} & \hc{\textbf{46.20}} \\
        \cmidrule{2-9}
        & MInference & 46.95\% & \textbf{54.91} & \textbf{50.36} & 46.93 & 42.62 & 34.05 & 45.77 \\
        & FlexPrefill (0.99) & 37.75\% & 52.77 & 48.91 & 46.39 & 43.06 & 36.60 & 45.55 \\
        & XAttention (0.95) & 47.31\% & 53.93 & 49.59 & 47.10 & 44.65 & 38.19 & 46.69 \\
        & \hc{\textbf{UniSparse (0.95)}} & \hc{47.51\%} & \hc{53.91} & \hc{49.96} & \hc{\textbf{47.76}} & \hc{\textbf{44.95}} & \hc{\textbf{38.58}} & \hc{\textbf{47.03}} \\
        \bottomrule
    \end{tabular}
    }
\end{table*}

\textbf{Comparison with Existing Methods.} Table~\ref{tab:complexity_comparison} provides a systematic comparison with existing dynamic methods, categorizing them by evaluation scope: \textbf{global} methods consider all query-key interactions, while \textbf{local} methods sample only a subset of queries to estimate importance.

\sysname achieves global evaluation by considering the full query-key interaction space in a compressed domain. Compared to local evaluation methods (MInference~\cite{jiang2024minference}, FlexPrefill~\cite{lai2025flexprefill}), which suffer from structural bias and modality-specific limitations as discussed in Section~\ref{sec:related_work}, \sysname provides broader coverage of semantic dependencies. Compared to the other global method XAttention~\cite{xu2025xattention}, \sysname achieves substantially lower selection overhead through multi-granularity compression. In fact, with appropriate compression factors $(c_q, c_k, c_h)$, \sysname's overhead can be comparable to local methods, while maintaining the superior accuracy benefits of global evaluation (validated in \S\ref{subsec:efficiency}). This positions \sysname at the favorable "sweet spot" of the proxy-performance trade-off identified in Section~\ref{sec:summary_gap}: structurally simple for low overhead, yet representationally robust for high-quality block selection across diverse modalities.

\section{Hardware-Aware Implementation}
\label{subsec:implementation}
\textbf{Library and Integration.} \sysname is a standalone sparse attention library exposing a simple \texttt{unisparse\_attn} API, that accepts $(\mathbf{Q}, \mathbf{K}, \mathbf{V})$ and returns $\mathbf{O}$. All complexity—including token selection, compression, and block-sparse computation—is encapsulated, enabling a drop-in replacement of standard attention layers. \sysname supports arbitrary compression factors $(c_q, c_k, c_h)$ with guaranteed causal correctness for auto-regressive models, and employs kernel templating and auto-tuning to enable aggressive thread-level parallelization at higher compression ratios, thereby optimizing GPU occupancy across configurations. This enables seamless integration with existing frameworks (e.g., PyTorch, Transformers~\cite{wolf-etal-2020-transformers}) without requiring model architecture changes or retraining. Combined with orthogonal techniques for decoding acceleration (e.g., KV cache compression and quantization~\cite{li2024snapkvllmknowslooking,tang2024questqueryawaresparsityefficient,li2025kvtunersensitivityawarelayerwisemixedprecision}), \sysname enables efficient end-to-end long-context inference.

\textbf{Fused Kernel Optimization.} Algorithm~\ref{alg:unisparse} involves multiple sequential operations (e.g., \texttt{compression}, \texttt{softmax}, \texttt{Top-P}) that would incur substantial memory traffic if implemented naively. Our fused kernel design keeps intermediate results in on-chip memory (shared memory and registers) rather than global HBM, combining these operations into unified kernel launches to minimize data movement. For causal masking, we precompute valid attention ranges and integrate masking directly into scoring kernels. These optimizations ensure $C_{\text{select}}$ remains a small fraction of end-to-end latency. For sparse attention computation, we integrate with block-sparse FlashAttention kernels that skip pruned blocks while maintaining memory efficiency.

\section{Experiments}
\label{sec:experiments}
This section evaluates \sysname across models and workloads and addresses three key questions:
(i) Can \sysname generalize across models and modalities?
(ii) Does \sysname achieve hardware efficiency?
(iii) How does each design choice contribute to the overall results?

\subsection{Experimental Setup}
\label{subsec:setup}
All experiments are run on a single NVIDIA H200 GPU. For fairness, all methods apply sparse attention computation only during the prefill stage, while maintaining full KV cache and dense computation during decode stage.

\textbf{Models.} 
We evaluate \sysname on three representative language and vision-language models to assess its generality across architectures. \emph{(1) Meta-Llama-3.1-8B-Instruct}~\cite{Llama-3.1-8B-Instruct} is a decoder-only model supporting 128K tokens via RoPE scaling. \emph{(2) Qwen2.5-7B-Instruct}~\cite{Qwen2.5-7B-Instruct} extends the context length to 128K using YaRN interpolation and provides strong multilingual capability. \emph{(3) Qwen2.5-VL-7B-Instruct}~\cite{Qwen2.5-VL-7B-Instruct} is a unified vision-language model that handles both text and video inputs. Together, these models cover diverse positional encoding schemes and modalities, enabling a comprehensive evaluation of \sysname.

\textbf{Workloads.}
To evaluate \sysname across diverse modalities and reasoning complexities—from synthetic to real-world tasks and from text to video understanding—we employ three challenging long-context benchmarks that together provide comprehensive coverage of long-context challenges across modalities and reasoning levels.
(1) RULER~\cite{hsieh2024ruler} is a synthetic benchmark with controllable task complexity, testing core long-context abilities such as retrieval, aggregation, and multi-hop reasoning over sequences up to 128K tokens. 
(2) HELMET~\cite{yen2025helmet} is a real-world benchmark spanning seven application-oriented categories (e.g., long-document QA, summarization, RAG) across diverse domains, requiring deep semantic understanding of real documents up to 128K tokens. 
(3) Video-MME~\cite{fu2024video} evaluates full-spectrum video understanding over 6 visual domains and 30 subfields, incorporating multi-modal inputs (frames, subtitles, audio) and testing temporal as well as cross-modal reasoning. 

\begin{table*}[!t]
    \centering
    \renewcommand{\arraystretch}{0.9}
    \caption{Performance comparison on Video-MME using Qwen2.5-VL-7B-Instruct. We report accuracy on short ($<$ 2 min), medium (4-15 min), and long ($>$ 30 min) videos, with and without subtitle inputs. }
    \label{tab:videomme}
    \setlength{\aboverulesep}{1.5pt}
    \setlength{\belowrulesep}{1.5pt}
    \setlength{\heavyrulewidth}{1.5pt}
    \resizebox{\textwidth}{!}{
    \begin{tabular}{l|c|c|c|c|c|c|c|c|c|c}
        \toprule
        \multirow{2}{*}{\textbf{Method}} & \multicolumn{5}{c|}{\textbf{Without Subtitles}} & \multicolumn{5}{c}{\textbf{With Subtitles}} \\
        \cline{2-11}
        & \textbf{Sparsity(↑)} & \textbf{Short} & \textbf{Medium} & \textbf{Long} & \textbf{Overall(↑)} & \textbf{Sparsity(↑)} & \textbf{Short} & \textbf{Medium} & \textbf{Long} & \textbf{Overall(↑)} \\
        \midrule
        FlashAttention & - & 75.1 & 66.7 & 54.2 & 65.3 & - & 75.3 & 71.7 & 61.7 & 69.6 \\
        \midrule
        FlexPrefill (0.95) & 57.9\% & 73.9 & 65.8 & \textbf{53.7} & 64.4 & 66.3\% & \textbf{75.2} & 69.9 & 61.7 & 68.9 \\
        XAttention (0.9) & 60.2\% & 73.6 & 63.6 & 51.9 & 63.0 & 61.1\% & 74.0 & 69.8 & 60.4 & 68.1 \\
        \highlightrow \textbf{UniSparse (0.9)} & 57.7\% & \textbf{74.4} & \textbf{65.9} & 53.3 & \textbf{64.6} & 58.6\% & 74.9 & \textbf{71.2} & \textbf{63.4} & \textbf{69.9} \\
        \midrule
        FlexPrefill (0.99) & 27.3\% & 74.6 & \textbf{66.8} & 53.7 & \textbf{65.0} & 36.3\% & \textbf{75.4} & \textbf{71.7} & 61.7 & 69.6 \\
        XAttention (0.95) & 49.2\% & 74.1 & 64.7 & 52.8 & 63.9 & 50.4\% & 75.2 & 70.8 & 61.1 & 69.0 \\
        \highlightrow \textbf{UniSparse (0.95)} & 45.9\% & \textbf{74.8} & 65.8 & \textbf{54.3} & \textbf{65.0} & 47.2\% & 75.3 & \textbf{71.7} & \textbf{62.0} & \textbf{69.7} \\
        \bottomrule
    \end{tabular}
    }
\vspace{-3mm}
\end{table*}

\textbf{Baselines and Configuration.} 
\sysname applies multi-granularity compression with global evaluation, using $c_q{=}c_k{=}8$ and Top-$P$ thresholds $P{=}0.9, 0.95$ by default; additional compression settings are analyzed in Section~\ref{subsec:ablation}. 
We compare \sysname against four widely adopted baselines under comparable sparsity levels, following official parameter recommendations denoted as $\tau$ and $\gamma$. 
(1) FlashAttention~\cite{dao2022flashattention} serves as the full attention baseline with memory-efficient implementation, providing the accuracy upper bound. 
(2) MInference~\cite{jiang2024minference} adopts offline pattern search and head-specific sparse configurations; we use its official searched files for each model. 
(3) FlexPrefill~\cite{lai2025flexprefill} employs a lightweight local evaluation using the last Q-block as probe, evaluated with $\tau{=}0.1$ and sparsity levels $\gamma{=}0.95, 0.99$. 
(4) XAttention~\cite{xu2025xattention} performs global evaluation via anti-diagonal scoring with strided sampling ($\text{stride}{=}8$), tested with $\tau{=}0.9, 0.95$. 
We organize comparisons into two sparsity groups: \textbf{moderate sparsity} (eg., FlexPrefill-0.99, XAttention-0.95) and \textbf{high sparsity} (eg., FlexPrefill-0.95, XAttention-0.9).

\begin{figure}[!t]  
  \centering
  \setlength{\abovecaptionskip}{-1pt}
  \setlength{\belowcaptionskip}{-6pt}
  \includegraphics[width=0.95\columnwidth]{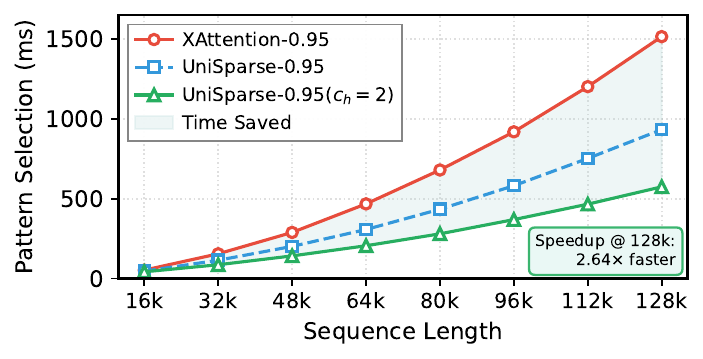}
  \caption{Block selection speedup across sequence lengths.}
  \label{fig:selection_overhead}
  \crunch
  \crunch
  \crunch
\end{figure}

\begin{figure*}[!t]  
  \centering
  \setlength{\abovecaptionskip}{-1pt}
  \setlength{\belowcaptionskip}{-2pt}
  \includegraphics[width=\textwidth]{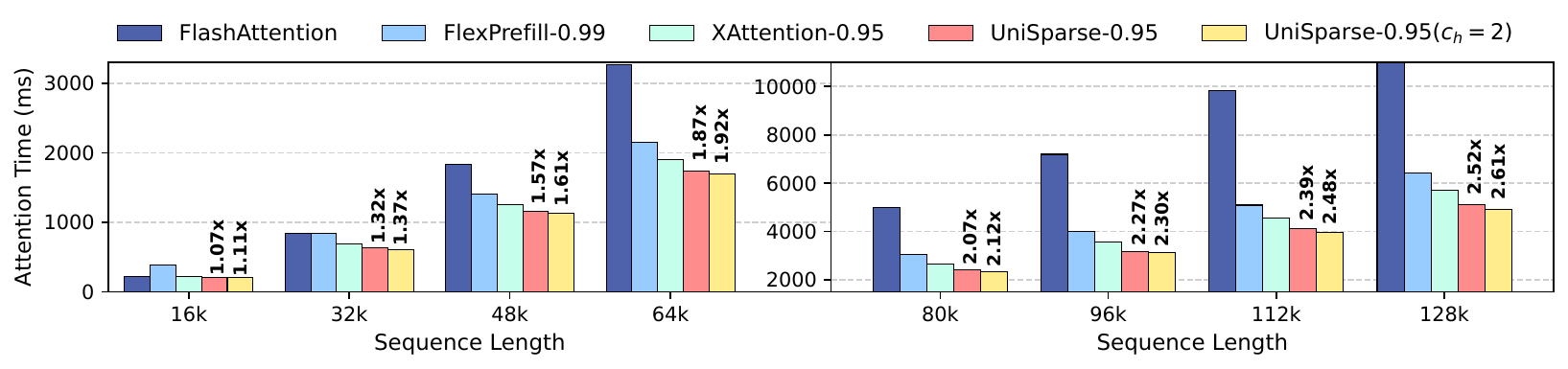}
  \caption{End-to-end attention speedup across sequence lengths (16K-128K tokens).}
  \label{fig:sdt}
  \vspace{-3mm}
\end{figure*}

\subsection{Main Results}
\label{subsec:main_results}
Tables~\ref{tab:ruler}--\ref{tab:videomme} summarize results on text benchmarks (RULER, HELMET) using Meta-Llama-3.1-8B-Instruct and Qwen2.5-7B-Instruct, and on the video benchmark (Video-MME) using Qwen2.5-VL-7B-Instruct. Across all modalities and context lengths (up to 128K tokens or long-duration videos), \sysname consistently achieves superior accuracy–efficiency trade-offs.

\textbf{Synthetic Long-Context Tasks.} On RULER's synthetic tasks, UniSparse achieves the best overall performance among sparse methods across both sparsity settings. On Llama, UniSparse-0.95 achieves 92.56 overall accuracy, matching FlexPrefill-0.99 at 92.59 while maintaining higher sparsity (46.33\% vs. 36.51\%). On Qwen, UniSparse-0.95 significantly outperforms all baselines, achieving 84.26 overall accuracy compared to MInference's 83.58 and XAttention-0.95's 83.63. Notably, UniSparse demonstrates consistent performance across all context lengths, from 4K to 128K tokens, with particularly strong results at extreme lengths where long-range dependency modeling is critical.

\textbf{Real-World Semantic Understanding.} The advantages of UniSparse become more pronounced on HELMET's real-world tasks, which require complex semantic understanding. On Llama, UniSparse-0.95 achieves 56.21 overall accuracy with 48.65\% sparsity, outperforming both FlexPrefill-0.99 at 56.01 and MInference at 55.37. Notably, UniSparse maintains consistent performance across different context lengths, demonstrating robust long-range dependency modeling. On Qwen, UniSparse-0.95 achieves 47.03 overall accuracy, surpassing all baselines including MInference's 45.77 and XAttention-0.95's 46.69. The consistent improvements across both models and benchmarks validate UniSparse's effectiveness in text-based long-context scenarios. Detailed task-level analysis is provided in attached supplemental material Appendix A.

\textbf{Multi-Modal Video Understanding.} UniSparse demonstrates strong cross-modal reasoning capabilities in both subtitle configurations. Under high sparsity, UniSparse-0.9 achieves 64.6 overall accuracy (57.7\% sparsity) without subtitles, outperforming FlexPrefill-0.95 at 64.4 and significantly surpassing XAttention-0.9 at 63.0. With subtitles, UniSparse-0.9 achieves 69.9 accuracy with 58.6\% sparsity, even surpassing FlashAttention's full attention performance, demonstrating its ability to preserve critical cross-modal information under aggressive compression. 
These results highlight UniSparse's capability to handle temporal reasoning and multi-modal inputs where important information is distributed across visual and textual modalities.

\textbf{Overall Performance Summary.} Across all three benchmarks spanning text and video modalities, UniSparse demonstrates exceptional accuracy retention while computing less than half of attention blocks. It consistently retains over 99\% of FlashAttention's accuracy and even surpasses full attention performance in certain cases. Compared to existing strong sparse attention methods, UniSparse's multi-granularity compression enables more effective adaptive evaluation that captures both coarse and fine-grained attention patterns. This results in consistent performance advantages across synthetic reasoning, real-world semantic understanding, and multi-modal video tasks without requiring task-specific tuning.

\begin{table*}[!t]
    \centering
    \renewcommand{\arraystretch}{0.95}
    \caption{Ablation study on compression strategies. Results show the average sparsity and overall scores computed over context lengths ranging from 4K to 128K for Meta-Llama-3.1-8B-Instruct and Qwen2.5-7B-Instruct models.
    }
    \label{tab:ablation_aggregation_strategy}
    \setlength{\aboverulesep}{1.5pt}
    \setlength{\belowrulesep}{1.5pt}
    \setlength{\heavyrulewidth}{1.5pt}
    \resizebox{\textwidth}{!}{
    \begin{tabular}{l|cccc|cccc}
        \toprule
        & \multicolumn{4}{c|}{\textbf{\textit{Meta-Llama-3.1-8B-Instruct}}} & \multicolumn{4}{c}{\textbf{\textit{Qwen2.5-7B-Instruct}}} \\
        \cmidrule(lr){2-5} \cmidrule(lr){6-9}
        \textbf{Method} & \textbf{Sparsity (↑)} & \textbf{RULER (↑)} & \textbf{Sparsity (↑)} & \textbf{HELMET (↑)} & \textbf{Sparsity (↑)} & \textbf{RULER (↑)} & \textbf{Sparsity (↑)} & \textbf{HELMET (↑)} \\
        \midrule
        \highlightrow \textbf{UniSparse} & 46.33\% & \textbf{92.56} & 48.65\% & \textbf{56.21} & 47.25\% & \textbf{84.26} & 47.51\% & \textbf{47.03} \\
        Max & 53.93\% & 92.11 & 56.85\% & 54.83 & 56.09\% & 81.01 & 58.98\% & 43.46 \\
        Stochastic & 51.68\% & 92.21 & 54.44\% & 55.23 & 54.21\% & 80.84 & 56.39\% & 44.05 \\
        \bottomrule
    \end{tabular}
    } 
\vspace{-3mm}
\end{table*}

\subsection{Efficiency Results}
\label{subsec:efficiency}
We measure efficiency on the HELMET benchmark using Meta-Llama-3.1-8B-Instruct, whose real-world tasks better reflect practical deployment scenarios.

\textbf{Block Selection Overhead.} Figure~\ref{fig:selection_overhead} compares the selection overhead $C_{\text{select}}$ of global evaluation methods across sequence lengths from 16K to 128K tokens. We exclude FlexPrefill and MInference due to their limited local evaluation scope as discussed in Table~\ref{tab:complexity_comparison}. UniSparse-0.95 achieves 1.62$\times$ speedup over XAttention-0.95 at 128K tokens. Unlike XAttention's fixed-stride sampling, UniSparse performs global evaluation incorporating information from all query and key tokens while reducing scoring complexity to $\mathcal{O}(L^2 h d_k / (c_q c_k c_h))$ through multi-granularity compression, ensuring comprehensive attention pattern capture with lower computational cost. With optional head compression ($c_h = 2$), UniSparse-0.95 achieves 2.64$\times$ speedup over XAttention-0.95, further reducing redundant cross-head evaluations.

\textbf{End-to-End Attention Speedup.} Figure~\ref{fig:sdt} presents the end-to-end attention computation time, including both block selection and sparse attention computation. At 128K tokens, UniSparse-0.95 achieves a 2.52$\times$ speedup over FlashAttention while retaining over 99\% of its accuracy. With head compression($c_h=2$), UniSparse reaches 2.61$\times$ speedup. We exclude MInference, a high-fidelity proxy method, due to its expensive offline pattern search. UniSparse consistently outperforms FlexPrefill and XAttention across all sequence lengths. FlexPrefill-0.99, despite lower selection overhead from local evaluation, is slower overall due to higher sparsity requirements caused by imprecise local block selection. XAttention-0.95 achieves competitive attention computation time but incurs higher selection overhead, leading to longer total latency. These results demonstrate that UniSparse's balanced design—efficient multi-granularity compression for selection combined with effective sparse pattern identification—delivers superior end-to-end efficiency while maintaining the accuracy advantages.

\begin{table*}[!t]
    \centering
    \renewcommand{\arraystretch}{0.9}
    \caption{Ablation study on Q-K compression ratio allocation. We fix the total compression factor ($c_q \times c_k = 64$) while varying the allocation between queries and keys.}
    \label{tab:ablation_qk_allocation}
    \setlength{\aboverulesep}{1pt}
    \setlength{\belowrulesep}{1pt}
    \setlength{\heavyrulewidth}{1.2pt}
    \resizebox{\textwidth}{!}{
    \begin{tabular}{l|cccc|cccc}
        \toprule
        & \multicolumn{4}{c|}{\textbf{\textit{Meta-Llama-3.1-8B-Instruct}}} & \multicolumn{4}{c}{\textbf{\textit{Qwen2.5-7B-Instruct}}} \\
        \cmidrule(lr){2-5} \cmidrule(lr){6-9}
        \textbf{Method} & \textbf{Sparsity (↑)} & \textbf{RULER (↑)} & \textbf{Sparsity (↑)} & \textbf{HELMET (↑)} & \textbf{Sparsity (↑)} & \textbf{RULER (↑)} & \textbf{Sparsity (↑)} & \textbf{HELMET (↑)} \\
        \midrule
        $c_q=32, c_k=2$ & 48.32\% & 91.37 & 49.75\% & 54.88 & 49.90\% & 81.93 & 49.94\% & 45.87 \\
        $c_q=16, c_k=4$ & 46.84\% & 92.46 & 49.02\% & 55.63 & 48.93\% & 83.61 & 49.07\% & 46.89 \\
        \highlightrow \textbf{UniSparse} & 46.33\% & 92.56 & 48.65\% & \textbf{56.21} & 47.25\% & \textbf{84.26} & 47.51\% & 47.03 \\
        $c_q=4, c_k=16$ & 46.32\% & 92.54 & 48.67\% & 55.99 & 47.48\% & 84.11 & 47.81\% & \textbf{47.27} \\
        $c_q=2, c_k=32$ & 44.88\% & \textbf{92.57} & 47.15\% & 55.97 & 45.68\% & 83.93 & 46.11\% & 47.24 \\
        \bottomrule
    \end{tabular}
    }
\end{table*}

\begin{table*}[!t]
    \centering
    \renewcommand{\arraystretch}{0.9}
    \caption{Ablation study on compression granularity. We vary the compression factor $c$ while maintaining balanced Q-K compression ($c_q = c_k = c$). Our default choice $c=8$ provides an effective balance between accuracy and efficiency.}
    \label{tab:ablation_compression_granularity}
    \setlength{\aboverulesep}{1.5pt}
    \setlength{\belowrulesep}{1.5pt}
    \setlength{\heavyrulewidth}{1.5pt}
    \resizebox{\textwidth}{!}{
    \begin{tabular}{l|cccc|cccc}
        \toprule
        & \multicolumn{4}{c|}{\textbf{\textit{Meta-Llama-3.1-8B-Instruct}}} & \multicolumn{4}{c}{\textbf{\textit{Qwen2.5-7B-Instruct}}} \\
        \cmidrule(lr){2-5} \cmidrule(lr){6-9}
        \textbf{Method} & \textbf{Sparsity (↑)} & \textbf{RULER (↑)} & \textbf{Sparsity (↑)} & \textbf{HELMET (↑)} & \textbf{Sparsity (↑)} & \textbf{RULER (↑)} & \textbf{Sparsity (↑)} & \textbf{HELMET (↑)} \\
        \midrule
        $c=4$ & 47.42\% & 92.59 & 49.79\% & 56.02 & 48.82\% & \textbf{84.51} & 49.09\% & 47.01 \\
        \highlightrow \textbf{UniSparse} & 46.33\% & 92.56 & 48.65\% & \textbf{56.21} & 47.25\% & 84.26 & 47.51\% & \textbf{47.03} \\
        $c=16$ & 46.86\% & \textbf{92.64} & 49.36\% & 55.94 & 48.84\% & 83.43 & 49.25\% & 46.54 \\
        $c=32$ & 46.16\% & 92.54 & 48.64\% & 55.87 & 48.30\% & 83.38 & 48.87\% & 46.49 \\
        \bottomrule
    \end{tabular}
    } 
\end{table*}

\begin{table*}[!t]
    \centering
    \renewcommand{\arraystretch}{0.9}
    \caption{Ablation study on head compression. We evaluate head-level compression factor $c_h$ under two Top-$P$ settings.}
    \label{tab:ablation_head_compression}
    \setlength{\aboverulesep}{1.5pt}
    \setlength{\belowrulesep}{1.5pt}
    \setlength{\heavyrulewidth}{1.5pt}
    \resizebox{\textwidth}{!}{
    \begin{tabular}{c|l|cccc|cccc}
        \toprule
        & & \multicolumn{4}{c|}{\textbf{\textit{Meta-Llama-3.1-8B-Instruct}}} & \multicolumn{4}{c}{\textbf{\textit{Qwen2.5-7B-Instruct}}} \\ \cmidrule(lr){3-6} \cmidrule(lr){7-10} \textbf{Top-$P$} & \textbf{Method} & \textbf{Sparsity (↑)} & \textbf{RULER (↑)} & \textbf{Sparsity (↑)} & \textbf{HELMET (↑)} & \textbf{Sparsity (↑)} & \textbf{RULER (↑)} & \textbf{Sparsity (↑)} & \textbf{HELMET (↑)} \\ \midrule
        \highlightrow 0.9 & \textbf{UniSparse} & 58.90\% & \textbf{92.44} & 61.82\% & 55.58 
        & 60.12\% & \textbf{83.07} & 60.84\% & 46.20 \\
        0.9 & $c_h=2$ & 57.54\% & 92.05 & 60.65\% & \textbf{55.72} & 53.90\% & 83.06 & 54.92\% & 46.53 \\
        0.9 & $c_h=4$ & 56.31\% & 91.82 & 59.64\% & 55.36 & 46.82\% & 82.74 & 47.94\% & \textbf{46.82} \\
        \midrule
        \highlightrow 0.95 & \textbf{UniSparse} & 46.33\% & \textbf{92.56} & 48.65\% & \textbf{56.21} 
        & 47.25\% & 84.26 & 47.51\% & 47.03 \\
        0.95 & $c_h=2$ & 45.13\% & 92.37 & 47.73\% & 56.02 & 41.73\% & \textbf{84.27} & 42.27\% & \textbf{47.27} \\
        0.95 & $c_h=4$ & 44.45\% & 92.30 & 47.39\% & 55.93 & 35.32\% & 83.99 & 36.18\% & 47.10 \\
        \bottomrule
    \end{tabular}
    }
\vspace{-3mm}
\end{table*}

\subsection{Ablation Study}
\label{subsec:ablation}
We conduct comprehensive ablation studies to analyze the impact of key design choices in UniSparse, including compression strategies, Q-K allocation ratios, compression granularity, and optional head compression.

\textbf{Compression Strategy.} Table~\ref{tab:ablation_aggregation_strategy} compares different compression strategies for constructing composite tokens. UniSparse adopts average pooling (Eq.~\ref{eq:avgq}) as the default strategy. While max pooling and stochastic pooling achieve higher sparsity ratios, they sacrifice accuracy due to incomplete information aggregation: max pooling only retains the most salient feature within each window, while stochastic pooling selects a single token probabilistically based on their normalized magnitudes, both failing to capture the full contextual information within the compression window. In contrast, average pooling preserves comprehensive information by uniformly aggregating all tokens within each window, enabling more accurate block importance estimation while maintaining competitive sparsity levels.

\textbf{Q-K Compression Ratio Allocation.} Table~\ref{tab:ablation_qk_allocation} investigates different allocations of compression ratios between queries and keys while fixing the total compression factor ($c_q \times c_k = 64$). Results show that balanced compression ($c_q = c_k = 8$) achieves the best overall performance across both models and benchmarks. Query-biased compression (e.g., $c_q=32, c_k=2$) leads to significant performance degradation, as over-compressing queries loses fine-grained distinctions needed for accurate importance scoring. Interestingly, key-biased compression (e.g., $c_q=4, c_k=16$) performs competitively in certain cases, particularly on Qwen2.5-7B, suggesting potential model-specific characteristics that warrant further investigation. Beyond accuracy considerations, balanced compression also enables more efficient implementations under causal masking, as symmetric Q-K granularity simplifies the computation of valid attention ranges during block selection.

\textbf{Compression Granularity.} Table~\ref{tab:ablation_compression_granularity} examines the impact of compression factor $c$ while maintaining balanced Q-K compression ($c_q=c_k=c$). Smaller compression factors (e.g., $c=4$) provide finer-grained representations that capture more nuanced attention patterns, but incurring higher selection overhead. Conversely, larger factors (e.g., $c=16$) reduce computational cost but risk over-smoothing local variations in attention importance, potentially missing critical fine-grained patterns. Our default choice of $c=8$ strikes a practical balance, achieving compression sufficient for efficient global evaluation while preserving enough granularity to distinguish important blocks, resulting in robust performance across diverse tasks and models.

\textbf{Head Compression.} Table~\ref{tab:ablation_head_compression} evaluates optional head-level compression under two sparsity settings. Without head compression ($c_h=1$, our default), UniSparse preserves complete cross-head information and achieves the best accuracy in most cases. Head compression leverages the observation that different heads often focus on similar important positions—compressing across heads distills this consensus while reducing selection overhead. However, head compression reduces achieved sparsity under the same Top-$P$ threshold because aggregating head information makes more blocks appear important (e.g., on Llama at Top-$P$=0.95, $c_h=2$ achieves 45.13\% vs. 46.33\% sparsity). 
The efficiency impact reflects this trade-off (\S\ref{subsec:efficiency}): while head compression dramatically reduces selection overhead (2.64$\times$ vs. 1.62$\times$ speedup over XAttention), end-to-end speedup is modest (2.61$\times$ vs. 2.52$\times$ over FlashAttention) due to lower sparsity. Overall, we provide head compression as an optional configuration that is most beneficial when selection overhead dominates or for tasks where attention patterns are more concentrated, with the accuracy-efficiency trade-off being task and model dependent.

\section{Conclusion}
In this work, we introduce \sysname, which leverages composite tokens as a universal summary for efficient sparse attention. 
Extensive evaluations across tasks and modalities show that \sysname achieves state-of-the-art accuracy while reducing computation costs. By bridging efficiency and generality, \sysname enables scalable long-context modeling and provides a practical foundation for future research in dynamic sparse attention.

\bibliography{example_paper}
\bibliographystyle{mlsys2025}
\clearpage
\appendix
\section{Detailed Analysis on HELMET Subtasks}
\label{appendix:helmet_subtasks}

Tables~\ref{tab:helmet_detailed_llama} and~\ref{tab:helmet_detailed_qwen} present detailed performance across HELMET's six subtasks, evaluating diverse aspects of long-context understanding from 8K to 128K tokens. We analyze each task category below.

\textbf{Retrieval-Augmented Generation (RAG).} This task evaluates the ability to identify and utilize relevant passages from retrieved documents. UniSparse-0.95 maintains competitive performance with FlashAttention across all sequence lengths on both models—achieving 59.21 vs. 58.67 at 128K on Llama and 46.88 vs. 46.75 at 128K on Qwen. The compressed global evaluation effectively captures document-level relevance patterns, consistently outperforming other sparse methods at comparable sparsity levels.

\textbf{Passage Re-Ranking (ReRank).} Re-ranking requires fine-grained discrimination between candidate passages. UniSparse-0.95 demonstrates strong performance at moderate lengths, particularly excelling at 16K-32K tokens where it achieves 52.32 and 40.80 on Llama versus XAttention-0.95's 50.21 and 40.04, and FlexPrefill-0.99's 50.31 and 43.53. On Qwen, UniSparse-0.95 leads consistently at longer sequences (26.26 at 32K, 14.25 at 64K, 8.34 at 128K), validating that Top-$P$ selection adapts effectively to fine-grained relevance signals.

\textbf{Generation with Citations (Cite).} Citation tasks demand precise localization of information sources. UniSparse-0.95 maintains competitive accuracy across sequence lengths, leading at 32K-64K on Qwen with scores of 13.59 and 12.45 respectively. While all sparse methods face challenges at extreme lengths due to the inherent difficulty of exact positional tracking under compression, UniSparse's global evaluation preserves sufficient positional information for accurate source attribution at practical sparsity levels.

\textbf{Long-Document QA (LongQA).} Long-form question answering requires comprehensive reasoning across extended contexts. UniSparse-0.95 excels on this task, consistently achieving top performance among sparse methods—reaching 46.54 versus XAttention-0.95's 45.60 at 128K on Llama, and 43.19 versus XAttention-0.95's 42.17 at 128K on Qwen. The advantage becomes more pronounced as context length increases, demonstrating that global evaluation scope enables effective capture of long-range dependencies essential for complex reasoning.

\textbf{Many-Shot In-Context Learning (ICL).} ICL evaluates the ability to leverage examples within context for novel tasks. UniSparse-0.95 maintains consistently high and stable performance across all sequence lengths on both models, progressing from 71.16 to 84.36 across 8K to 128K on Llama, and from 69.52 to 79.88 on Qwen, closely matching FlashAttention's trajectory. This demonstrates that average pooling preserves semantic relationships between in-context examples and queries, enabling effective pattern recognition under high sparsity without degradation at extreme lengths.

\textbf{Synthetic Recall (Recall).} Recall tests precise information retrieval from long contexts. UniSparse shows strong performance particularly where other methods degrade significantly. UniSparse-0.9 achieves perfect 100.00 at 8K on Llama, and maintains robust recall at extreme lengths with 95.25 at 128K on Llama and 47.81 at 128K on Qwen, substantially outperforming XAttention-0.95's 91.56 and 45.56 respectively. The comprehensive global evaluation ensures critical information is not missed, unlike strided sampling that creates blind spots.

\textbf{Summary.} Across all six task categories and sequence lengths, UniSparse consistently demonstrates superior or competitive performance compared to existing sparse attention methods while maintaining higher sparsity levels. The multi-granularity compression enables efficient global evaluation that adapts to diverse attention patterns—from fine-grained re-ranking to long-range reasoning—without task-specific tuning. Notably, UniSparse's advantage becomes more pronounced at extreme sequence lengths, where it maintains stable performance while other methods experience significant degradation. Combined with its superior computational efficiency (§\ref{subsec:efficiency})—achieving up to 2.61$\times$ speedup over FlashAttention and consistently outperforming other sparse methods in both selection overhead and end-to-end latency—this consistent effectiveness validates our core insight: compression-based ranking preserves the semantic fidelity necessary for accurate block selection across multiple modalities and heterogeneous real-world tasks, establishing UniSparse as a unified and robust solution for long-context understanding.

\begin{table*}[!htbp]
    \centering
    \caption{Detailed HELMET task breakdown for \textbf{Meta-Llama-3.1-8B-Instruct}. Each section shows performance on six sub-tasks (RAG, ReRank, Cite, LongQA, ICL, Recall) at a specific sequence length. The best results among sparse methods at similar sparsity levels are highlighted.}
    \label{tab:helmet_detailed_llama}
    \setlength{\aboverulesep}{1.5pt}
    \setlength{\belowrulesep}{1.5pt}
    \setlength{\heavyrulewidth}{1.5pt}
    \resizebox{0.8\textwidth}{!}{
    \begin{tabular}{l|l|cccccc|c}
        \toprule
        \textbf{Seqlen} & \textbf{Method} & \textbf{RAG} & \textbf{ReRank} & \textbf{Cite} & \textbf{LongQA} & \textbf{ICL} & \textbf{Recall} & \textbf{Overall (↑)} \\
        \midrule
        \multirow{8}{*}{\textbf{8K}} 
        & FlashAttention & 70.08 & 65.74 & 32.65 & 26.89 & 71.00 & 100.00 & 61.06 \\
        \cmidrule{2-9}
        & FlexPrefill (0.95) & 65.62 & 53.43 & 22.61 & 28.45 & 68.52 & 96.69 & 55.89 \\
        & XAttention (0.9) & 68.67 & \textbf{66.03} & 29.21 & 26.54 & 70.48 & 97.94 & 59.81 \\
        & \hc{\textbf{UniSparse (0.9)}} & \hc{\textbf{69.67}} & \hc{64.53} & \hc{\textbf{30.84}} & \hc{\textbf{29.38}} & \hc{\textbf{70.72}} & \hc{\textbf{100.00}} & \hc{\textbf{60.86}} \\
        \cmidrule{2-9}
        & MInference & \textbf{69.67} & \textbf{66.59} & \textbf{34.37} & 27.08 & 70.92 & \textbf{100.00} & \textbf{61.44} \\
        & FlexPrefill (0.99) & 68.96 & 62.91 & 33.52 & \textbf{28.26} & 71.00 & \textbf{100.00} & 60.66 \\
        & XAttention (0.95) & 69.00 & 66.08 & 30.94 & \textbf{28.26} & 71.04 & 99.25 & 60.76 \\
        & \hc{\textbf{UniSparse (0.95)}} & \hc{69.58} & \hc{66.25} & \hc{33.84} & \hc{27.40} & \hc{\textbf{71.16}} & \hc{\textbf{100.00}} & \hc{61.37} \\
        \midrule
        \multirow{8}{*}{\textbf{16K}} 
        & FlashAttention & 67.29 & 51.18 & 24.85 & 33.59 & 75.72 & 99.75 & 58.73 \\
        \cmidrule{2-9}
        & FlexPrefill (0.95) & 65.08 & 41.86 & 14.58 & \textbf{35.16} & 75.44 & 96.25 & 54.73 \\
        & XAttention (0.9) & 67.29 & 48.57 & \textbf{20.08} & 34.21 & 76.24 & 98.00 & 57.40 \\
        & \hc{\textbf{UniSparse (0.9)}} & \hc{\textbf{67.38}} & \hc{\textbf{49.39}} & \hc{18.60} & \hc{34.87} & \hc{\textbf{76.48}} & \hc{\textbf{99.50}} & \hc{\textbf{57.70}} \\
        \cmidrule{2-9}
        & MInference & 67.33 & 50.28 & 21.17 & 32.74 & 76.24 & \textbf{100.00} & 57.96 \\
        & FlexPrefill (0.99) & 67.33 & 50.31 & \textbf{21.47} & \textbf{33.64} & 75.56 & \textbf{100.00} & 58.05 \\
        & XAttention (0.95) & 67.33 & 50.21 & 20.70 & 33.30 & 76.28 & 99.00 & 57.80 \\
        & \hc{\textbf{UniSparse (0.95)}} & \hc{\textbf{67.50}} & \hc{\textbf{52.32}} & \hc{20.78} & \hc{33.45} & \hc{\textbf{76.52}} & \hc{99.50} & \hc{\textbf{58.34}} \\
        \midrule
        \multirow{8}{*}{\textbf{32K}} 
        & FlashAttention & 65.21 & 43.01 & 10.43 & 42.66 & 78.44 & 99.69 & 56.57 \\
        \cmidrule{2-9}
        & FlexPrefill (0.95) & 64.58 & 30.38 & 8.35 & 39.28 & \textbf{79.16} & 97.06 & 53.14 \\
        & XAttention (0.9) & \textbf{65.33} & 39.62 & 10.48 & 41.20 & 78.48 & 98.19 & 55.55 \\
        & \hc{\textbf{UniSparse (0.9)}} & \hc{65.29} & \hc{\textbf{40.80}} & \hc{\textbf{10.83}} & \hc{\textbf{41.78}} & \hc{79.08} & \hc{\textbf{99.50}} & \hc{\textbf{56.21}} \\
        \cmidrule{2-9}
        & MInference & 64.92 & 40.31 & \textbf{11.59} & 37.65 & \textbf{79.36} & 99.12 & 55.49 \\
        & FlexPrefill (0.99) & \textbf{65.58} & \textbf{43.53} & 9.28 & \textbf{41.98} & 77.60 & 99.38 & \textbf{56.22} \\
        & XAttention (0.95) & 65.08 & 40.04 & 10.91 & 41.43 & 78.80 & 99.25 & 55.92 \\
        & \hc{\textbf{UniSparse (0.95)}} & \hc{65.12} & \hc{40.80} & \hc{10.22} & \hc{40.70} & \hc{78.84} & \hc{\textbf{99.50}} & \hc{55.87} \\
        \midrule
        \multirow{8}{*}{\textbf{64K}} 
        & FlashAttention & 64.71 & 33.06 & 11.32 & 45.65 & 81.56 & 99.19 & 55.92 \\
        \cmidrule{2-9}
        & FlexPrefill (0.95) & 63.58 & 16.22 & 6.15 & 40.36 & 81.92 & 94.44 & 50.45 \\
        & XAttention (0.9) & 64.33 & 28.46 & 6.00 & \textbf{45.01} & \textbf{82.44} & 96.19 & 53.74 \\
        & \hc{\textbf{UniSparse (0.9)}} & \hc{\textbf{64.50}} & \hc{\textbf{29.80}} & \hc{\textbf{6.67}} & \hc{44.41} & \hc{82.36} & \hc{\textbf{97.69}} & \hc{\textbf{54.24}} \\
        \cmidrule{2-9}
        & MInference & 62.21 & 31.02 & 8.52 & 38.59 & \textbf{82.88} & 98.94 & 53.69 \\
        & FlexPrefill (0.99) & \textbf{64.96} & \textbf{32.97} & \textbf{10.46} & 43.86 & 80.92 & \textbf{99.50} & \textbf{55.44} \\
        & XAttention (0.95) & 64.83 & 29.57 & 9.17 & 44.62 & 82.00 & 98.31 & 54.75 \\
        & \hc{\textbf{UniSparse (0.95)}} & \hc{64.58} & \hc{32.41} & \hc{9.50} & \hc{\textbf{44.81}} & \hc{82.04} & \hc{98.31} & \hc{55.27} \\
        \midrule
        \multirow{8}{*}{\textbf{128K}} 
        & FlashAttention & 58.67 & 14.23 & 2.76 & 44.30 & 83.92 & 94.88 & 49.79 \\
        \cmidrule{2-9}
        & FlexPrefill (0.95) & \textbf{59.88} & 5.19 & 2.54 & 44.32 & 85.40 & \textbf{93.19} & 48.42 \\
        & XAttention (0.9) & 58.62 & 9.60 & \textbf{3.03} & 43.96 & 85.32 & 85.88 & 47.74 \\
        & \hc{\textbf{UniSparse (0.9)}} & \hc{59.38} & \hc{\textbf{9.87}} & \hc{2.78} & \hc{\textbf{44.52}} & \hc{\textbf{85.44}} & \hc{91.50} & \hc{\textbf{48.91}} \\
        \cmidrule{2-9}
        & MInference & 55.88 & \textbf{14.35} & 1.72 & 40.58 & 83.84 & 93.38 & 48.29 \\
        & FlexPrefill (0.99) & 58.62 & 13.73 & 2.88 & 44.09 & 83.28 & \textbf{95.50} & 49.68 \\
        & XAttention (0.95) & 59.17 & 12.46 & 2.74 & 45.60 & \textbf{84.68} & 91.56 & 49.37 \\
        & \hc{\textbf{UniSparse (0.95)}} & \hc{\textbf{59.21}} & \hc{12.76} & \hc{\textbf{3.23}} & \hc{\textbf{46.54}} & \hc{84.36} & \hc{95.25} & \hc{\textbf{50.22}} \\
        \bottomrule
    \end{tabular}
    }
\end{table*}

\begin{table*}[!htbp]
    \centering
    \caption{Detailed HELMET task breakdown for \textbf{Qwen2.5-7B-Instruct}. Each section shows performance on six sub-tasks (RAG, ReRank, Cite, LongQA, ICL, Recall) at a specific sequence length.}
    \label{tab:helmet_detailed_qwen}
    \setlength{\aboverulesep}{1.5pt}
    \setlength{\belowrulesep}{1.5pt}
    \setlength{\heavyrulewidth}{1.5pt}
    \resizebox{0.8\textwidth}{!}{
    \begin{tabular}{l|l|cccccc|c}
        \toprule
        \textbf{Seqlen} & \textbf{Method} & \textbf{RAG} & \textbf{ReRank} & \textbf{Cite} & \textbf{LongQA} & \textbf{ICL} & \textbf{Recall} & \textbf{Overall (↑)} \\
        \midrule
        \multirow{8}{*}{\textbf{8K}} 
        & FlashAttention & 60.83 & 58.77 & 27.61 & 34.23 & 69.56 & 79.62 & 55.10 \\
        \cmidrule{2-9}
        & FlexPrefill (0.95) & 52.33 & 39.94 & 12.69 & 32.32 & 67.92 & 60.19 & 44.23 \\
        & XAttention (0.9) & 59.38 & \textbf{57.09} & \textbf{25.37} & 30.96 & \textbf{69.40} & 72.56 & 52.46 \\
        & \hc{\textbf{UniSparse (0.9)}} & \hc{\textbf{59.92}} & \hc{56.45} & \hc{24.51} & \hc{\textbf{33.34}} & \hc{69.24} & \hc{\textbf{75.31}} & \hc{\textbf{53.13}} \\
        \cmidrule{2-9}
        & MInference & \textbf{60.79} & \textbf{59.04} & 26.07 & 32.28 & \textbf{69.92} & \textbf{81.38} & \textbf{54.91} \\
        & FlexPrefill (0.99) & 58.46 & 51.87 & 24.09 & \textbf{35.91} & \textbf{69.92} & 76.38 & 52.77 \\
        & XAttention (0.95) & 59.58 & 57.73 & \textbf{27.54} & 32.05 & 69.64 & 77.06 & 53.93 \\
        & \hc{\textbf{UniSparse (0.95)}} & \hc{60.21} & \hc{56.57} & \hc{26.36} & \hc{31.75} & \hc{69.52} & \hc{79.06} & \hc{53.91} \\
        \midrule
        \multirow{8}{*}{\textbf{16K}} 
        & FlashAttention & 59.25 & 37.70 & 16.77 & 37.35 & 74.40 & 77.75 & 50.54 \\
        \cmidrule{2-9}
        & FlexPrefill (0.95) & 50.67 & 22.92 & 7.83 & 35.25 & 73.32 & 54.75 & 40.79 \\
        & XAttention (0.9) & \textbf{59.88} & \textbf{37.46} & \textbf{13.18} & \textbf{37.41} & \textbf{74.80} & 65.19 & 47.99 \\
        & \hc{\textbf{UniSparse (0.9)}} & \hc{58.96} & \hc{37.30} & \hc{13.08} & \hc{37.27} & \hc{\textbf{74.80}} & \hc{\textbf{70.31}} & \hc{\textbf{48.62}} \\
        \cmidrule{2-9}
        & MInference & 58.38 & 37.54 & 14.87 & 38.69 & 73.84 & \textbf{78.88} & \textbf{50.36} \\
        & FlexPrefill (0.99) & 56.88 & 36.26 & 13.72 & \textbf{38.89} & 73.36 & 74.38 & 48.91 \\
        & XAttention (0.95) & \textbf{60.08} & 38.32 & 15.03 & 36.95 & \textbf{74.40} & 72.75 & 49.59 \\
        & \hc{\textbf{UniSparse (0.95)}} & \hc{59.42} & \hc{\textbf{38.48}} & \hc{\textbf{15.46}} & \hc{37.34} & \hc{74.36} & \hc{74.69} & \hc{49.96} \\
        \midrule
        \multirow{8}{*}{\textbf{32K}} 
        & FlashAttention & 56.21 & 25.79 & 13.87 & 42.51 & 76.56 & 76.12 & 48.51 \\
        \cmidrule{2-9}
        & FlexPrefill (0.95) & 48.75 & 13.43 & 6.73 & 37.44 & \textbf{76.52} & 56.38 & 39.87 \\
        & XAttention (0.9) & 54.88 & 20.05 & \textbf{13.20} & \textbf{43.79} & 76.28 & 65.56 & 45.63 \\
        & \hc{\textbf{UniSparse (0.9)}} & \hc{\textbf{55.75}} & \hc{\textbf{25.26}} & \hc{11.98} & \hc{41.43} & \hc{76.40} & \hc{\textbf{70.75}} & \hc{\textbf{46.93}} \\
        \cmidrule{2-9}
        & MInference & 53.96 & 23.96 & 11.13 & 41.55 & 74.16 & \textbf{76.81} & 46.93 \\
        & FlexPrefill (0.99) & 55.33 & 22.55 & 12.57 & 38.60 & 74.64 & 74.62 & 46.39 \\
        & XAttention (0.95) & \textbf{55.88} & 25.75 & 13.26 & \textbf{42.76} & \textbf{76.52} & 68.44 & 47.10 \\
        & \hc{\textbf{UniSparse (0.95)}} & \hc{55.71} & \hc{\textbf{26.26}} & \hc{\textbf{13.59}} & \hc{41.90} & \hc{76.40} & \hc{72.69} & \hc{\textbf{47.76}} \\
        \midrule
        \multirow{8}{*}{\textbf{64K}} 
        & FlashAttention & 55.21 & 14.49 & 10.93 & 44.67 & 77.00 & 66.81 & 44.85 \\
        \cmidrule{2-9}
        & FlexPrefill (0.95) & 43.92 & 6.83 & 4.07 & 35.17 & 76.56 & 51.12 & 36.28 \\
        & XAttention (0.9) & \textbf{55.67} & \textbf{15.85} & 8.49 & \textbf{47.03} & \textbf{77.24} & 54.75 & 43.17 \\
        & \hc{\textbf{UniSparse (0.9)}} & \hc{55.12} & \hc{13.83} & \hc{\textbf{10.59}} & \hc{46.88} & \hc{77.20} & \hc{\textbf{61.69}} & \hc{\textbf{44.22}} \\
        \cmidrule{2-9}
        & MInference & 51.83 & 15.46 & 8.54 & 41.85 & 70.56 & \textbf{67.50} & 42.62 \\
        & FlexPrefill (0.99) & 53.83 & 13.58 & 8.12 & 43.45 & 75.68 & 63.69 & 43.06 \\
        & XAttention (0.95) & \textbf{55.67} & \textbf{15.78} & 11.81 & \textbf{47.09} & \textbf{77.08} & 60.50 & 44.65 \\
        & \hc{\textbf{UniSparse (0.95)}} & \hc{54.96} & \hc{14.25} & \hc{\textbf{12.45}} & \hc{46.94} & \hc{76.88} & \hc{64.19} & \hc{\textbf{44.95}} \\
        \midrule
        \multirow{8}{*}{\textbf{128K}} 
        & FlashAttention & 46.75 & 7.76 & 6.57 & 45.68 & 79.28 & 49.94 & 39.33 \\
        \cmidrule{2-9}
        & FlexPrefill (0.95) & 34.12 & 3.75 & 2.48 & 24.07 & 77.48 & 38.44 & 30.06 \\
        & XAttention (0.9) & \textbf{47.33} & 6.44 & 4.76 & 40.71 & \textbf{80.04} & 42.25 & 36.92 \\
        & \hc{\textbf{UniSparse (0.9)}} & \hc{47.29} & \hc{\textbf{7.24}} & \hc{\textbf{5.54}} & \hc{\textbf{40.76}} & \hc{\textbf{80.04}} & \hc{\textbf{47.81}} & \hc{\textbf{38.11}} \\
        \cmidrule{2-9}
        & MInference & 44.54 & 5.34 & 6.29 & 38.15 & 69.28 & 40.69 & 34.05 \\
        & FlexPrefill (0.99) & 44.46 & 7.39 & 5.37 & 36.56 & 78.20 & 47.62 & 36.60 \\
        & XAttention (0.95) & \textbf{47.67} & 8.01 & \textbf{6.31} & 42.17 & 79.44 & 45.56 & 38.19 \\
        & \hc{\textbf{UniSparse (0.95)}} & \hc{46.88} & \hc{\textbf{8.34}} & \hc{5.36} & \hc{\textbf{43.19}} & \hc{\textbf{79.88}} & \hc{\textbf{47.81}} & \hc{\textbf{38.58}} \\
        \bottomrule
    \end{tabular}
    }
\end{table*}

\end{document}